\documentclass[a4,twocolumn]{article}

\usepackage[round]{natbib}
\usepackage[utf8]{inputenc}
\usepackage{amsmath,amsthm,amssymb}
\usepackage{algorithm,algorithmic}
\usepackage{graphicx,pgfgantt}
\usepackage{booktabs,url,pifont}
\usepackage{caption}
\usepackage{subcaption}

\definecolor{linkblue}{rgb}{0.10,0.40,0.70}

\usepackage[colorlinks=true,citecolor=linkblue]{hyperref}

\def\f{\mathbf{f}}
\def\x{\mathbf{x}}
\def\tx{\tilde{\mathbf{x}}}

\def\z{\mathbf{z}}

\def\1{\mathbf{1}}
\def\0{\mathbf{0}}
\def\R{\mathbb{R}}

\def\L{\mathcal{L}}
\def\y{\mathbf{y}}
\def\B{\mathbf{B}}

\def\U{\mathbf{U}}
\def\W{\mathbf{W}}
\def\s{\mathbf{s}}

\def\Z{\mathbf{Z}}

\def\bmu{\boldsymbol\mu}
\def\bs{\boldsymbol\sigma}

\def\0{\mathbf{0}}
\def\N{\mathcal{N}}
\def\D{\mathcal{D}}
\def\b{\mathbf{b}}
\def\E{\mathbb{E}}

\def\diag{\mathrm{diag}}

\def\kl{\mathrm{KL}}

\DeclareMathOperator*{\argmax}{arg\,max}
\DeclareMathOperator*{\argmin}{arg\,min}

\usepackage[margin=0.9in]{geometry}

\renewenvironment{abstract}
 {
  \begin{center}
  \bfseries \abstractname\vspace{-.5em}\vspace{0pt}
  \end{center}
  \list{}{
    \setlength{\leftmargin}{.8cm}%
    \setlength{\rightmargin}{\leftmargin}%
  }%
  \item\relax}
 {\endlist}

\begin{document}

%\runningtitle{short title}
%\runningauthor{Trinh, Kaski, Heinonen}

\title{\Large{\bf{Scalable Bayesian neural networks by layer-wise input augmentation}}}
\author{Trung Trinh$^1$ $\qquad\qquad$ Samuel Kaski$^{1,2}$ $\qquad\qquad$ Markus Heinonen$^1$\vspace{7pt}
\\
$^1$Aalto University, Finland \\ $^2$University of Manchester, UK} 
\date{}

\maketitle

%\twocolumn[
%\aistatstitle{Scalable Bayesian neural networks by layer-wise input augmentation}
%\aistatsauthor{Trung Trinh$^1$ \And Samuel Kaski$^{1,2}$ \And Markus Heinonen$^1$}
%\aistatsaddress{$^1$Aalto University \And $^2$University of Manchester} 
%]

%
% relevant papers collected from Wilson2020 and FBNN paper. 'x' marks cited papers.
%
%%% discussion papers
% x Wilson20  "case for deep learning"
% Seeger06  "Bayesian modelling review"
% x Zhang18   "Understanding deep learning by rethinking generalisation"
% x Ovadia19  "Can you trust your model uncertainty?"
% Neal96    "Bayesian learning for neural networks"
% Kendall17 "What uncertainties we need in BDL"
% x Gustafson19 "Evaluating scalable BNNs"
% x Guo17     "On calibration of DNNs"
% Gelman13  "BDA"
% Wenzel20 "How good is Bayesian posterior?"

%%% Bayesian papers
%
% x Pradier18, Latent parameter BNNs
% Ritter18, Laplace DNN
% Graves11, variational BNN (old)
% x Blundel15, VI-BNN
% Saatchi17, Bayes-GAN
% Wilson15, DKL, DNN + GP
% x Sun19, FBNN, DNN + GP
% x Gal16, MC-dropout
% Kingma15, variational dropout
% x Depeweg17, LV-BNN
% Ghosh18, horseshoe BNN
% Lobato15, probabilistic backprop for BNN
% Louizos17, norm flows for VI-BNNs
% x Ma17, implicit processes
% Sun17, structured BNNs
% Zhang18, noisy natural gradients
%
%%% DNN papers
%
% x Lakshminarayanan17, deep ensembles
% x Garipov18, fast ensembling
% x Khan18, weight perturbations
% x Izmailov18, SWA
% x Maddox19, SWAG
% Zhang20, SWA-MCMC
% Hafner18, noise contrastive priors
%

\begin{abstract}
We introduce implicit Bayesian neural networks, a simple and scalable approach for uncertainty representation in deep learning. Standard Bayesian approach to deep learning requires the impractical inference of the posterior distribution over millions of parameters. Instead, we propose to induce a distribution that captures the uncertainty over neural networks by augmenting each layer's inputs with latent variables. We present appropriate input distributions and demonstrate state-of-the-art performance in terms of calibration, robustness and uncertainty characterisation over large-scale, multi-million parameter image classification tasks.
\end{abstract}

\section{INTRODUCTION}

Deep neural networks (DNNs) excel at learning remarkably complex patterns from large datasets in many machine learning tasks and domains. However, DNNs are typically overconfident in their predictions on unseen or noisy data \citep{zhang2017,wilson2020}, due to optimizing the parameter vector to represent only a single high performance function hypothesis of the data. Learning neural networks with confidence calibration \citep{guo2017}, robustness to perturbations \citep{su2019}, or accurate uncertainty characterisation \citep{ovadia2019,gustafsson2020} still face many major challenges, which require careful -- and time-consuming -- architectural tuning.

% heuristic approaches to fix calibration + others
Multiple approaches have been proposed to alleviate these issues. In stochastic weight averaging (SWA), neural parameters are averaged over stochastic gradient descent iterates to improve generalization \citep{izmailov2018} and the point estimate is further expanded into a local distribution of solutions \citep{maddox2019simple}. In deep ensembles a collection of alternative neural networks are learned \citep{lakshminarayanan2017simple,garipov18} from different initialisations to cover multiple data hypotheses. Dropout has been repurposed to carry out uncertainty estimation \citep{gal2016dropout}. 

\begin{table}[b!]
    \caption{Common network architectures with number of weight (incl. biases) or node parameters to infer, obtained from \texttt{torchvision.models}. Here we define a node as either an input feature for fully-connected layers or an input channel for convolution layers.}
    \begin{center}
    \resizebox{\columnwidth}{!}{
    \begin{tabular}{lrrrr}
        \toprule
                &         &   \multicolumn{3}{c}{Parameters}  \\
                \cmidrule(lr){3-5}
        Network & Layers & weights & nodes & w/n ratio \\
        \midrule
        LeNet & 5 & 42K & 23 & 1800x \\
        AlexNet & 8 & 61M & 18,307 & 3300x \\
        VGG16-small & 16 & 15M & 5,251 & 2900x \\
        VGG16-large & 16 & 138M & 36,995 & 3700x \\
        ResNet50 & 50 & 26M & 24,579 & 1000x \\
        WideResNet-28x10 & 28 & 36M & 9,475 & 3800x \\
        \bottomrule
    \end{tabular}
    }
    \end{center}
    \label{tab:networks}
\end{table}

% BNN fixes
Bayesian neural networks (BNNs) offer a principled approach of learning a posterior distribution of neural networks compatible with both data and parameter priors \citep{blundell2015weight}. By marginalizing the prediction over a large set of hypotheses, we can model the epistemic uncertainty, that is, which of the typically overparameterised networks are plausible, as a function of data \citep{wilson2020}. The impractically high computational cost of full Bayesian inference has necessitated posterior approximations, for instance SG-MCMC \citep{wenzel2020good}, variational inference \citep{blundell2015weight,louizos2017,khan2018}, expectation propagation \citep{zhao2020}, and moment matching \citep{gal2016dropout}. BNNs have been extended with more expressive functional priors \citep{sun2019functional}, and with latent variables to capture unobserved stochasticity \citep{depeweg2017uncertainty}. Despite these advances, BNNs are still not practical for modern network architectures, such as  AlexNet, VGG-16 or WideResNet-28x10, due to the high number of parameters in these models.

We argue that inferring posteriors over millions of parameters is inherently unrealistic, ultimately leading to the unscalability of BNNs to large networks \citep{wenzel2020good}. Recently, the pioneering work of \citet{pradier2018} proposed to avoid the weight-space posterior pathology by learning a lower rank parameter representation.

%A line of method combines Gaussian processes, a family of nonparametric function distributions, with convolution tasks \citep{wilk17,heinonen19} by placing a process priors on the neural network 

In this paper we introduce one of the first Bayesian neural network models that can be applied to architectures with over 10 millions weights. Our key contribution is to infer posteriors over input nodes instead of weights. Here we define an input node as either an input feature for fully-connected layers or an input channel for convolution layers. In typical neural architectures, this can potentially reduce the dimension of the posterior by up to 3800 fold. (See Table \ref{tab:networks}).

Our contributions include:
\begin{itemize}
    \item Introducing orders of magnitude smaller input posteriors for neural networks compared to weight posteriors.
    \item Introducing efficient Bayesian neural networks for architectures with over 10 million parameters.
    \item State-of-the-art Bayesian neural networks in calibration and accuracy for large-scale image classification tasks.
    \item We release PyTorch implementation at \url{github.com/trungtrinh44/ibnn}, with support for using pre-trained models.
\end{itemize}

\section{IMPLICITLY BAYESIAN NEURAL NETWORK}

We consider the supervised learning task of inferring a neural network from a dataset of $N$ observations $\D = \{ \x_i, \y_i\}_{i=1}^N$ of multivariate inputs $\x \in \R^D$ and outputs $\y \in \R^P$, where the input can also represent images. 

In classic feed-forward $\sigma$-activated BNN $\sigma(\W \x + \b)$ a prior $p(\W,\b)$ is placed over stochastic weights $\W$ and bias $\b$, whose posterior $p(\W,\b | \D)$ is inferred \citep{blundell2015weight}. This can be interpreted as an implicit neural process with weights and biases as latent variables \citep{ma2019}. In this paper we propose instead to shift the posterior to \emph{layer-wise inputs} by introducing latent input variables, while keeping the weights and biases deterministic and subject to conventional optimisation. We denote this approach as \emph{implicitly} Bayesian neural network (iBNN), since the posterior over weights is implicitly induced from the explicit input priors.

The iBNN is defined
\begin{align}
    \f_{\ell}(\x) &= \sigma_\ell\big( \U_\ell (\z_\ell \circ \f_{\ell-1}) + \b_\ell\big), \quad \ell \geq 1 \\
    \f_0 &:= \x \\
    \z_\ell &\sim p(\z),
\end{align}
where the inputs of the $\ell$ layer are multiplied elementwise (denoted $\circ$) by latent random variables $\z_\ell$, while throughout this paper we consider $\U_\ell$ and $\b_\ell$ as ordinary, non-stochastic weights  and biases, which are both optimized. The latent variables $\z_\ell$ then induces an implicit process over the output nodes $\f_\ell$ \citep{ma2019}.

We note that our approach differs both from latent variable BNN's \citep{depeweg2017uncertainty} and the implicit processes \citep{ma2019}, which both explicitly infer the potentially very high-dimensional weight posterior.

\subsection{Layer-wise input priors}

\begin{figure*}[t]
    \centering
    \includegraphics[width=.99\textwidth]{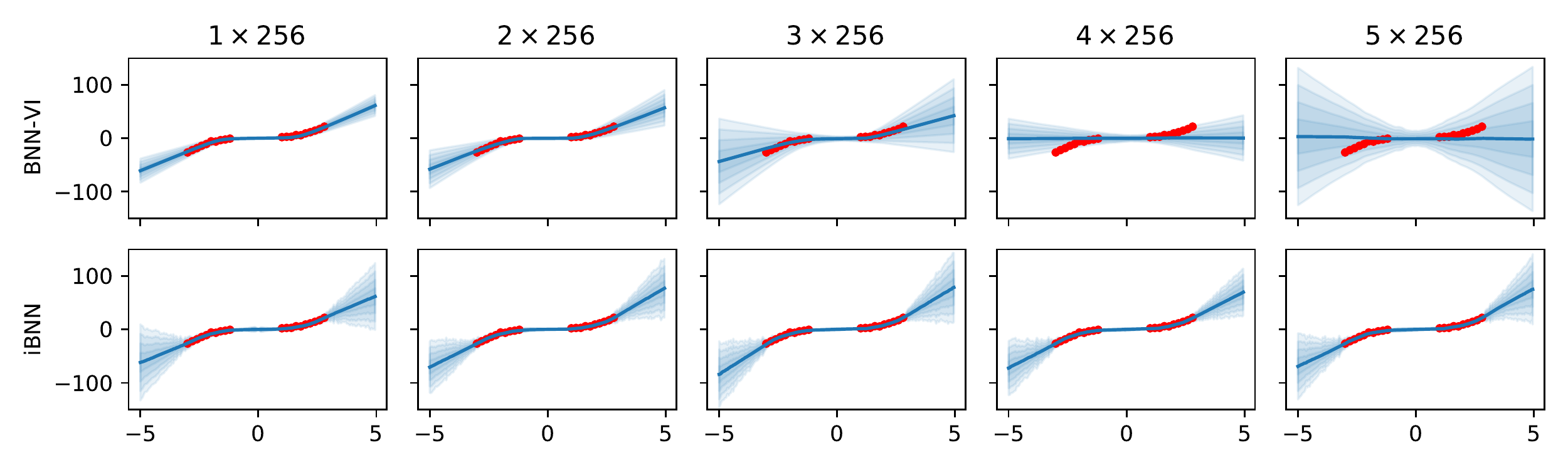}
    \caption{Behaviour of the BNN-VI and the iBNN on samples generated from a univariate function $y = x^3$ as we increase the number of hidden layers from $1$ to $5$ with $256$ nodes per layer. We visualise the dataset (red points) and predictive posterior (blue) over the input x-axis and output y-axis.}
    \label{fig:toy_regress}
\end{figure*}

In this section, we explain our motivation of using multiplicative noise by showing a connection between the iBNN and the conventional BNN. We begin by considering a single-layer iBNN 
\begin{align}
    \y &= \sigma(\U \tx + \b)  \\
    \tx &\sim p(\tx|\x),
\end{align}
where $\y \in \R^{k}$ is the output, and $\U \in \R^{k \times D}$ is the non-stochastic weight-matrix, $\b \in \R^k$ is the bias vector, $\tx \in \R^D$ is the perturbed input, and $\sigma$ is the activation function. 

To find a suitable form of $p(\tx|\x)$, we compare this model with a single-layer classic BNN
\begin{align}
    \y &= \sigma(\W \x + \b) \\
    \W &\sim p(\W),
\end{align}
where $\W \in \R^{k \times D}$ is the stochastic weight matrix and $p(\W)$ is weight distribution. For these two models to be equivalent, we let each pair of corresponding summands in the matrix multiplications equal each other
\begin{align} \label{eq:eq}
    w_{m,n} x_n = u_{m,n} \tilde{x}_n,
\end{align}
where $m \in [1,k], n \in [1, d]$. The activations and biases can be ignored. When $x_n \neq 0$, we can represent the stochastic weight $w_{m,n}$ via the deterministic weight $u_{m,n}$ and the stochastic input $\tilde{x}_n$:
\begin{align}
    w_{m, n} &= u_{m,n} z_n \label{eq:wuz1} \\
    z_n &= \frac{\tilde{x}_n}{x_n}.
\end{align}
In the equation above, we define $z_n$ as the stochastic noise, which when multiplied by the deterministic weight $u_{m,n}$ gives us the distribution of the stochastic weight $w_{m,n}$. In other words, the distribution of the stochastic weight $\W$ is now defined as:
\begin{align}
    \W &= \U (\diag\,\z) \label{eq:WUz} \\
    \z &\sim p(\z),
\end{align}
which is the result of perturbing the input $\x$ with the same multiplicative noise $\z$:
\begin{align}
    \Tilde{\x} &= \x \circ \z \\
    \z &\sim p(\z).
\end{align}
% \begin{equation}
%     W \sim \U \diag Z
%     Z \sim p(\Z)
% \end{equation}
% \begin{equation}
%     W \sim p(W)
% \end{equation}

Eq. \eqref{eq:wuz1} shows that the same noise distribution is shared between weights that interact with the same input feature. For a convolution layer, this notion translates to the same noise distribution being shared between weights that interact with the same input filter. This perturbation is intuitive, in that the variance of $p(\Tilde{x}_n)$ depends on the magnitude of $x_n$, i.e., input features that provide stronger signal will also admit a wider range of sample values.

\subsection{Variational inference}

We aim to infer the `input' posterior $p(\Z | \D;\theta)$ over the $L$ noise variables $\Z = \z_1, \ldots, \z_L$, while treating the neural parameters $\theta = (\U,\B)$, that is, weights $\U = \U_1, \ldots, \U_L$ and biases $\B = \b_1, \ldots, \b_L$ as hyperparameters to be optimised. We infer a factorised variational posterior approximation $q_\phi(\z_{1:L}) = q_\phi(\z_1) \cdots q_\phi(\z_L)$ that is closest to the true (intractable) posterior in the KL sense,
\begin{align}
    \argmin_{\phi, \theta} \kl[ q_\phi(\Z) || p(\Z | \D; \theta)],
\end{align}
where $\phi$ are the variational parameters and $\theta$ are the neural weights, which are jointly optimised. 

This translates into maximizing the evidence lower bound (ELBO) \citep{blei2017}:
\begin{multline}
    \argmax_{\phi,\theta} \L = \E_{q_\phi(\z_1) \cdots q_\phi(\z_L)}\left[ \log p(\D | \z_{1:L}, \theta\right] \\
     \quad - \beta \sum_{\ell=1}^L  \kl\left[ q_\phi(\z_\ell) || p(\z_\ell) \right], \label{eq:kl}
\end{multline}
which balances the optimisation of the expected likelihood $p(\D | \Z,\theta)$ over the variational posterior $q_\phi(\z_{1:L})$, and the layer-wise KL prior terms between the variational posterior and prior $p(\z_\ell)$. We consider the $\beta$-weighted KL term for more flexible approximation \citep{alemi2018}. At $\beta=1$, the ELBO is a lower bound for the log evidence;
\begin{align}
    \log p(\D; \theta) = \log \int p(\D | \Z; \theta) p(\Z) d\Z \ge \L_{\beta=1},
\end{align}
while $\beta \not =1$ adjusts the balance between the likelihood and the prior. We consider deterministic weights and biases $(\U,\B)$, which can be either optimised or initialised directly from pre-trained models.

\subsection{Variational ensemble posterior}

We consider the mixture of Gaussians family of variational posterior distribution for the latent variables $\{ \z_\ell\}_{\ell=1}^L$ across layers $\ell = 1,\ldots,L$:
\begin{align} \label{eq:qz}
    q_\phi(\Z) &= \prod_{\ell=1}^L q_\phi(\z_\ell) \\
    q_\phi(\z_\ell) &= \frac{1}{K} \prod_{k=1}^K  \N\big(\z_{\ell,k} | \bmu_{\ell,k}, \diag \, \bs_{\ell,k}^2\big),
\end{align}
where $K$ is the number of mixture components, and the variational parameters $\phi = \{\bmu_{\ell,k}, \bs_{\ell,k}\}_{\ell=1,k=1}^{L,K}$ are their mean and standard deviations. Given the multiplicative noise, we assume Gaussian priors with mean $1$ and standard deviations $\s$ as hyperparmeter,
\begin{align}
    p(\Z) &= \prod_{\ell=1}^L p(\z_\ell) \\
    p(\z_\ell) &= \N\big( \z_\ell | \1, \diag \, \s^2\big).
\end{align}

\begin{algorithm}[t]
\caption{Training procedure}
\label{algo:training}
\begin{algorithmic}[1]
\REQUIRE $B$: batch size, $N$: number of epochs, $K$: number of components of the posterior, $S$: number of samples per component, $\sigma_0$: the initial standard deviation for posterior mean initialization.
\STATE Initialize the means of the components of the posterior using $\N(1, \sigma_0^2)$.
\FOR{$i=1$ \TO $N$}
    \FORALL{mini-batches of size $B$ from the dataset}
        \STATE Evenly assign $B/K$ data points to each of the $K$ components.
        \STATE For each data point, draw $S$ noise samples from its assigned component and calculate the loss according to Eq. \eqref{eq:kl}
        \STATE Update the model's parameters.
    \ENDFOR
\ENDFOR
\end{algorithmic}
\end{algorithm}

However, there is no closed form equation to calculate the KL divergence in Eq. \eqref{eq:kl} between the mixture $q(\z_\ell)$ and the Gaussian prior $p(\z_\ell)$. We propose a tractable variant of the mixture of Gaussians approximation by calculating the KL divergence over the mixture mean $\hat{q}(\z_\ell)$,
\begin{align}
    \hat{q}(\z_\ell) &= \N\big(\hat{\bmu}_\ell, \diag \, \hat{\bs}_\ell^2\big) \\
    \hat{\bmu}_\ell &= \frac{1}{K} \sum_{k=1}^K  \bmu_{\ell,k} \\
    \hat{\bs}_\ell^2 &= \frac{1}{K^2} \sum_{k=1}^K \bs_{\ell,k}^2,
\end{align}
which is the distribution of the random variable $\hat{\z}_\ell = \frac{1}{K} \sum_{k=1}^K \z_{\ell,k}$, where $\z_{\ell,k}$ is the sample drawn from the $k$-th component of the original $q(\z_\ell)$. 

\begin{figure}[t]
    \centering
    \begin{subfigure}[b]{0.511\columnwidth}
         \centering\includegraphics[width=\textwidth]{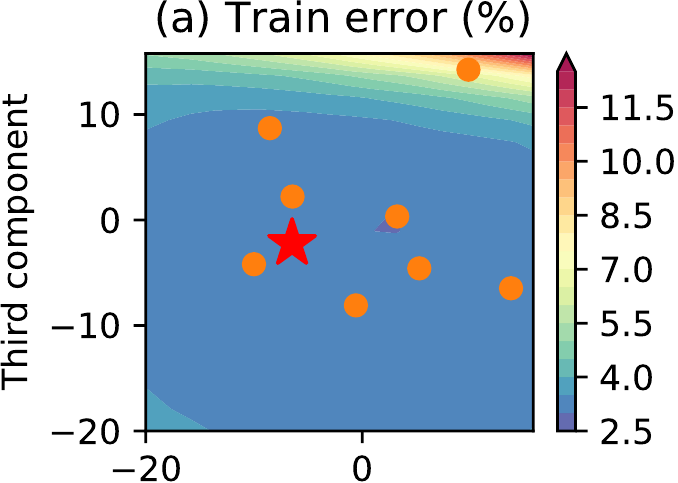}
     \end{subfigure}
     \hfill
     \begin{subfigure}[b]{0.466\columnwidth}
         \centering\includegraphics[width=\textwidth]{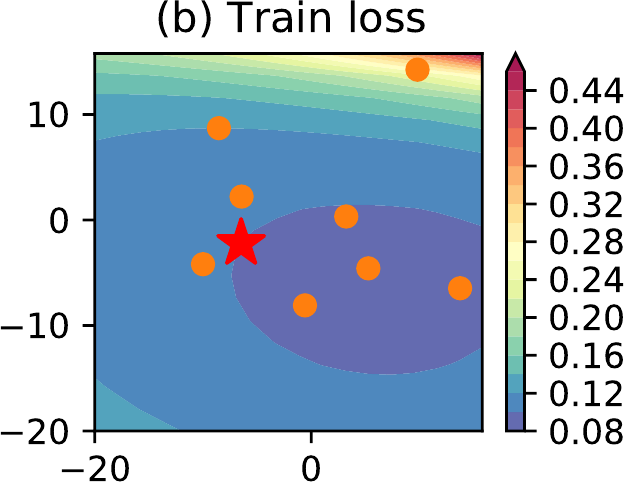}
    \end{subfigure}
    \\
    \begin{subfigure}[b]{0.511\columnwidth}
         \centering\includegraphics[width=\textwidth]{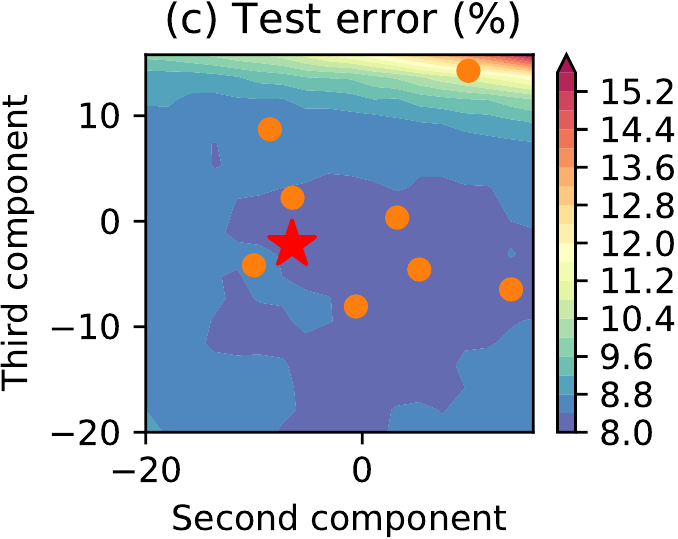}
     \end{subfigure}
     \hfill
     \begin{subfigure}[b]{0.466\columnwidth}
         \centering\includegraphics[width=\textwidth]{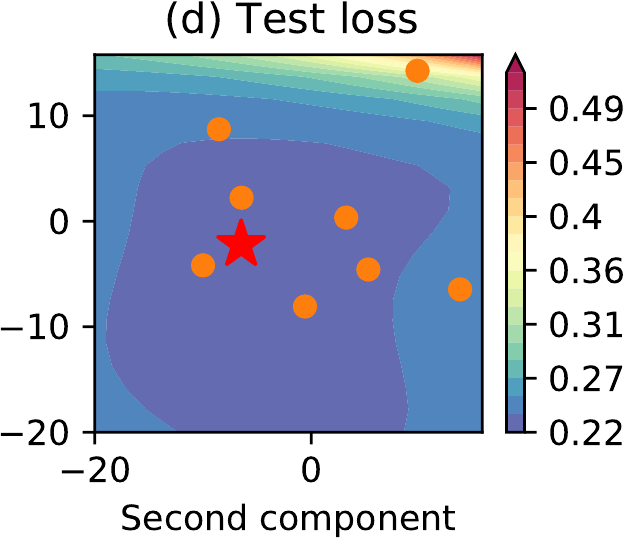}
     \end{subfigure}
    \caption{The PCA embedding of $K=8$ variational ensemble of LeNet on Fashion-MNIST shows the spread of the component-wise weights $\W_{(k)}$ (orange dots) along with the optimized weights $\U$ (red star). The iBNN has populated a wide local mode with its components, which leads to better generalization, as shown in plot (c) and (d).}
    \label{fig:pca}
\end{figure}

The KL term encourages the mean of the components to center around $\1$, which implies that they center around the layer-wise inputs $\f_{\ell-1}$, and the variances to center around $\s^2$. We optimize each component on their own stochastic minibatch slice, which leads to each component learning a different representation of the data (See Algorithm \ref{algo:training}). This \emph{ensemble} inference has two potential effects. First, through equation $\W = \U (\diag\,\z)$ \eqref{eq:WUz}, the $K$ components of the posterior of the latent variable $\z$ propagate into $K$ components of the stochastic (implicit) weights $\W$ which center around deterministic weights $\U$. This center-forcing effect encourages the deterministic weight $\U$ to converge to a wide optima on the loss surface, because it is surrounded by other high performing parameters. Second it allows the components of the posterior of the weight $\W$ to cover a broad region on the loss surface, which can capture different modes of the posterior, since there are no penalty on how far the components reside from the mean \citep{izmailov2018} (See Figure \ref{fig:pca}).

\begin{figure}[t]
    \centering
    \includegraphics[width=\columnwidth]{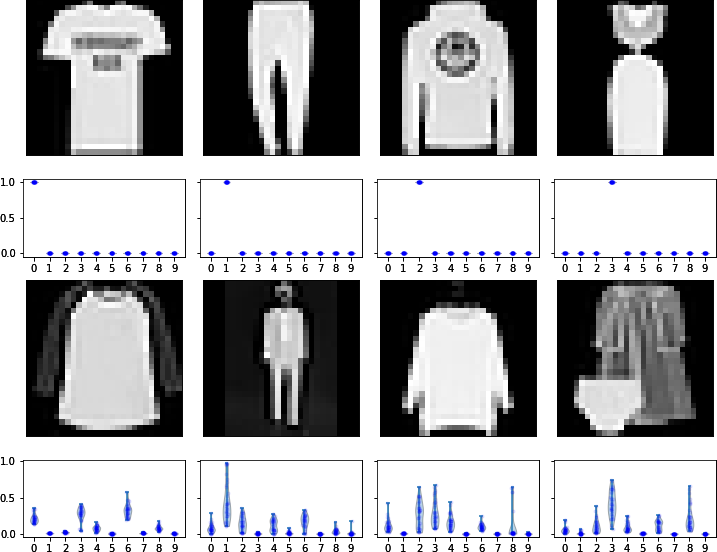}
    \caption{We visualize example softmax outputs of the iBNN for the first 4 classes in the Fashion-MNIST dataset. Each column represents a class. The top row contains samples with lowest predictive entropy and the bottom row contains those with highest predictive entropy.}
    \label{fig:fmnist_softmax}
\end{figure}

\subsection{Uncertainty decomposition}

We get the predictive distribution of the iBNN by marginalizing over the latent variables $\Z$:
\begin{equation}
    p(\y | \U, \x ) = \int p(\y|\U,\Z,\x) p(\Z) \mathrm{d}\Z
\end{equation}
Following \cite{depeweg2017uncertainty}, we can represent the total uncertainty using the predictive entropy $\mathrm{H}(\y|\U, \x)$. The aleatoric uncertainty is defined as
\begin{align} \label{eq:ale}
    \mathbb{E}_{q(\Z)}\left[\mathrm{H}(\y|\U,\Z,\x) \right],
\end{align} 
while the epistemic uncertainty is quantified by the mutual information between $\y$ and $\Z$,
\begin{equation} \label{eq:epi}
    \mathrm{MI}(\y, \Z) = \mathrm{H}(\y|\U, \x) - \mathbb{E}_{q(\Z)}\left[\mathrm{H}(\y|\U,\Z,\x) \right].
\end{equation}

\section{EXPERIMENTS}

In this section we will compare the proposed iBNN against the main competing methods of deterministic DNNs, BNN-VI \citep{blundell2015weight}, and MC-dropout \citep{gal2016dropout} on \textsc{cifar-10} and \textsc{cifar-100} datasets over large-scale VGG-16 and WideResNet-28x10 networks. We will demonstrate the iBNN's state-of-the-art performance in terms of predictive accuracy, robustness to model complexity, calibration, and out-of-distribution performance.

\subsection{1D regression}

We first demonstrate the iBNN against BNN-VI on a simple univariate regression problem of inferring the function $y=x^3$ with 30 datapoints. We use a a simple feed forward neural network with ReLU activation and 256 hidden nodes. We visualize in Figure \ref{fig:toy_regress} the behaviour of the iBNN and compare it to the behaviour of the BNN as we increase the number of hidden layers from $L=1,\ldots,5$. As the model size increases the BNN-VI collapses, as previously reported by \citet{su2019}. In contrast, the iBNN retains a good fit with good uncertainty representation independent of the model complexity. 

\subsection{Model complexity comparison}

Next, we compare the robustness of the iBNN and BNN-VI models on the relatively simple image classification dataset Fashion-MNIST using the LeNet architecture. Figure \ref{fig:fmnist_softmax} visualises the softmax outputs of the iBNN model, which shows that the model returns diverse predictions when it is highly uncertain.

\begin{figure}[t]
    \centering
    \includegraphics[width=.75\columnwidth]{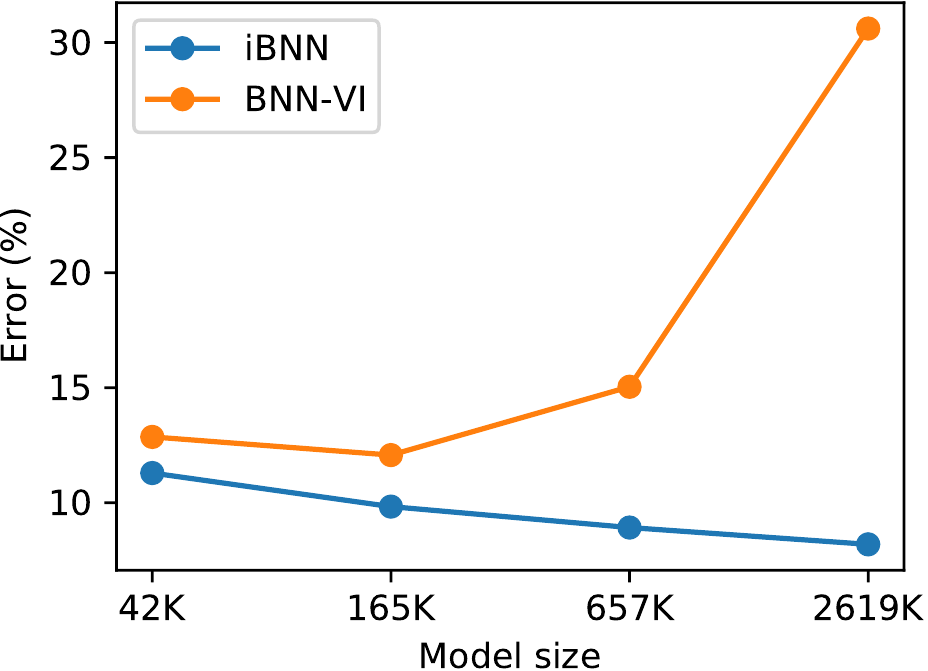}
    \caption{Error of iBNN and BNN-VI on the Fashion-MNIST dataset over LeNet architecture as number of parameters (weights and biases) in the neural network increases from 1x to \{4x,16x,64x\}, while number of layers do not change.} %. \red{change to error}}
    \label{fig:error_vs_size}
\end{figure}

Figure \ref{fig:error_vs_size} shows the error of the two models when we vary the number of parameters in LeNet by varying the number of hidden nodes. The iBNN improves its performance as network grows more complex, while BNN starts to degrade significantly after 165k parameters. The experiment demonstrates the pathology of variational inference over BNNs on scaling to larger networks. 

\subsection{Benchmarks}

\begin{table}[h]
\centering
\caption{Hyperparameters for each experiments. $\lambda_0$ is the initial learning rate for the deterministic weight $\U$, $\lambda_1$ is the constant learning rate for the parameters of the latent posterior.}
\label{tab:hyperparams}
\resizebox{\columnwidth}{!}{%
\begin{tabular}{lcccc}
\toprule
                    & \multicolumn{2}{c}{VGG-16}          & \multicolumn{2}{c}{WideResNet-28x10} \\ 
                    \cmidrule(lr){2-3} \cmidrule(lr){4-5}
             & \textsc{cifar-10}         & \textsc{cifar-100}        & \textsc{cifar-10}          & \textsc{cifar-100}        \\ \midrule
$\lambda_0$         & $0.05$           & $0.05$           & $0.1$             & $0.1$            \\
$\lambda_1$         & $1.2$            & $1.6$            & $2.4$             & $4.8$            \\
\# samples $S$                 & $2$              & $2$              & $2$               & $2$            \\
$\bmu_{\ell,k}$ init & $\N(1.0,0.75^2)$ & $\N(1.0,0.75^2)$ & $\N(1.0,0.5^2)$   & $\N(1.0,0.5^2)$  \\
Prior $p(\z_\ell)$  & $\N(1.0,0.3^2)$  & $\N(1.0,0.3^2)$  & $\N(1.0,0.1^2)$   & $\N(1.0,0.1^2)$  \\
Weight decay        & $0.0005$         & $0.0003$         & $0.0005$          & $0.0005$         \\ \bottomrule
\end{tabular}%
}
\end{table}

\paragraph{Experimental protocol.} We describe our training procedure in Algorithm \ref{algo:training}. We focus on evaluating iBNN on image classification tasks using two network architectures: VGG-16 \citep{vgg} and WideResNet-28x10 \citep{wrn} on the \textsc{cifar-10} and \textsc{cifar-100} datasets. 

\begin{figure*}[t]
    \centering
    \includegraphics[width=.8\textwidth]{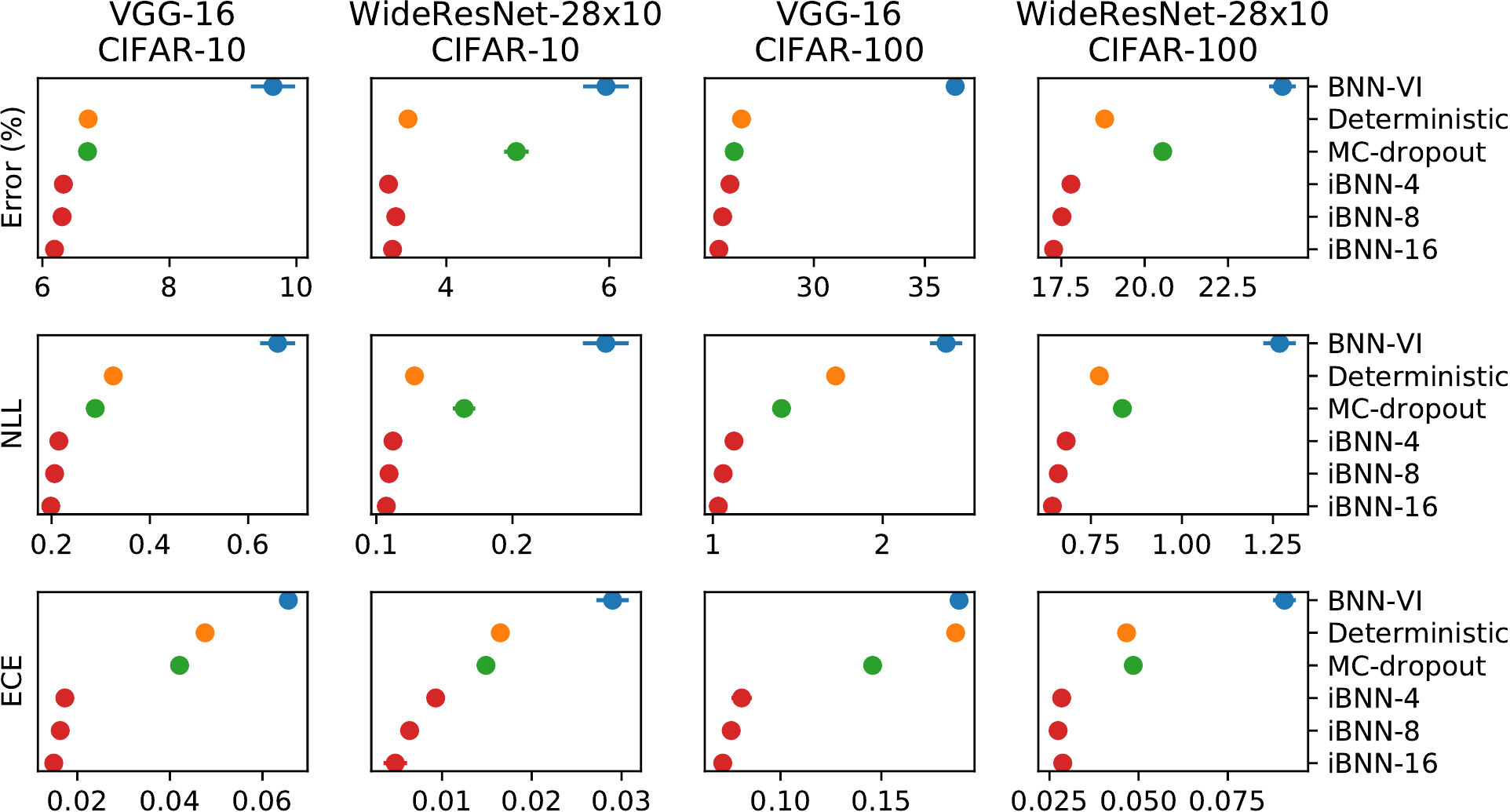}
    \caption{The state-of-the-art iBNN performance in image classification.  We visualise the test error (\%), NLL and ECE of the VGG-16 and WideResNet-28x10 models on the \textsc{cifar} datasets. Each setting is run 5 times with different random seeds. Standard deviations are shown with bars, most of which are too small to visualise (See Supplements for corresponding numerical table). Lower is better in all quantities.}
    \label{fig:benchmark}
\end{figure*}

We run each experiment for 300 epochs using a batch size of 128. We use SGD with Nesterov momentum of 0.9 as our optimizer. For the deterministic weight, we initially set the learning rate to $\lambda_0$, then starting to anneal the learning rate linearly to $0.01\lambda_0$ during 120 epochs starting from epoch 150. For the posterior parameters of the latent variables, we use a constant large learning rate $\lambda_1$ throughout training, which we set to a large value so that these latent variables can quickly adapt to the changes of the deterministic weights. We also linearly anneal the weight of the KL term $\beta$ from $0$ to $1$ during the first 200 epochs. We initialize the mean of each component in the latent posterior using $\N(1, \sigma_0^2)$, where $\sigma_0$ is a hyperparameter, and the corresponding standard deviation is initialized using $\N(0.05,0.02^2)$.

\begin{figure}[t]
    \centering
    \includegraphics[width=\columnwidth]{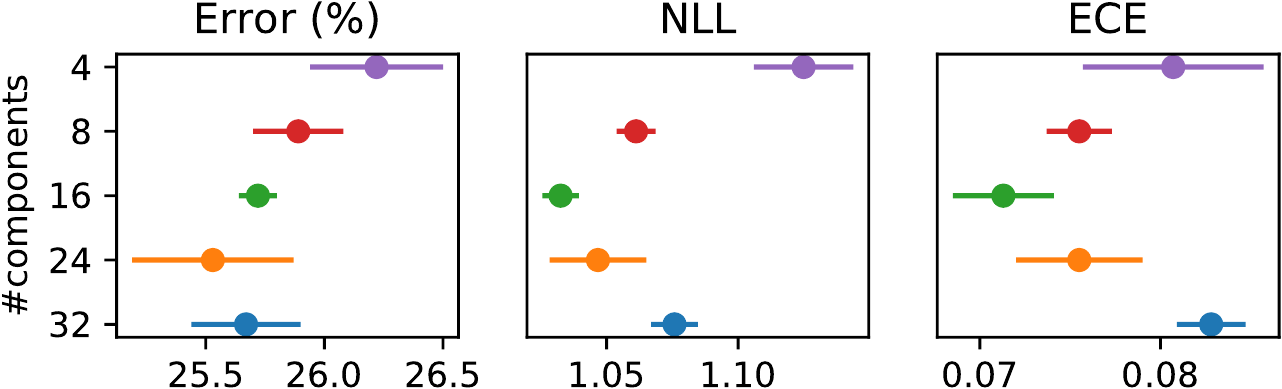}
    \caption{Performance of VGG-16 model on \textsc{cifar-100} with different number of components $K \in \{4,8,16,24,32\}$ in the variational posterior $q(\z_\ell)$ \eqref{eq:qz}. Lower is better in all quantities.}
    \label{fig:effect_component}
\end{figure}

For this method, we find four hyperparameters that are crucial to the final performance: the number of components $K$, the standard deviation $\sigma_0$ of the normal distribution for posterior mean initialization, the standard deviation $s$ of the prior distribution, and the learning rate of the variational parameters $\lambda_1$. We find the best value for the latter three hyperparameters for $K \in \{4, 8, 16\}$ using $10\%$ of the training data as validation, and then report the results using those hyperparameters after training on full datasets. These hyperparameters are reported in Table \ref{tab:hyperparams}.  We find that it is crucial to set a high learning rate $\lambda_1$ for the variational parameters so that the latent variables can quickly adapt to the changes of the deterministic weight during training, and to set an appropriate standard deviation $s$ for the prior to control the 'strictness` of the KL term.

\paragraph{Evaluation criteria.} We compare iBNN against DNNs, 
MC-dropout \citep{gal2016dropout} and BNN-VI \citep{blundell2015weight} on three metrics: classification error, negative log-likelihood (NLL) and expected calibration error (ECE) \citep{guo2017}. To evaluate the performance of the iBNN, we draw 5 samples from each of the $K$ components of the posterior, which gives us a total of $5 \times K$ sample predictions, which are combined to form the final prediction. We report the performance of the iBNN with 4, 8 and 16 components in the posterior. The results are shown in Figure \ref{fig:benchmark}.

% Report the ECE error, calibration plots
\begin{figure*}[t]
    \centering
    \begin{subfigure}[b]{0.224\textwidth}
         \centering
         \includegraphics[width=\textwidth]{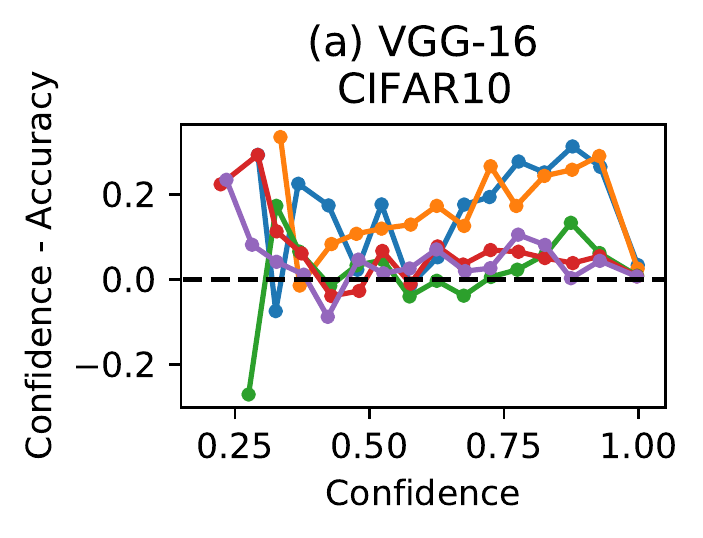}
     \end{subfigure}
     \hfill
     \begin{subfigure}[b]{0.211\textwidth}
         \centering
         \includegraphics[width=\textwidth]{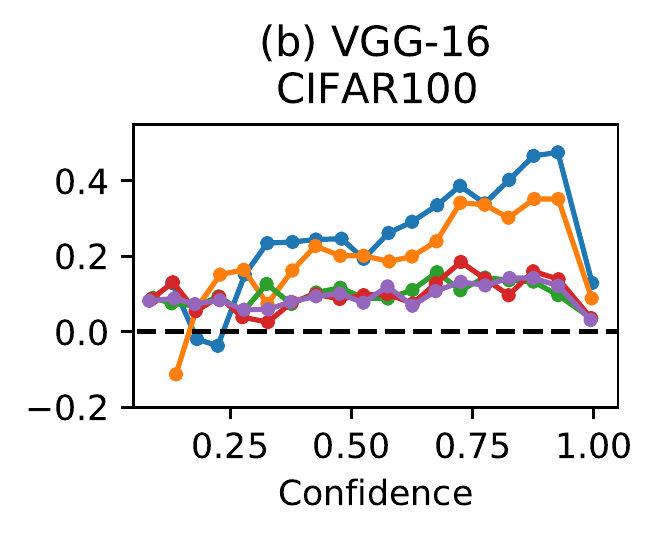}
     \end{subfigure}
     \hfill
     \begin{subfigure}[b]{0.211\textwidth}
         \centering
         \includegraphics[width=\textwidth]{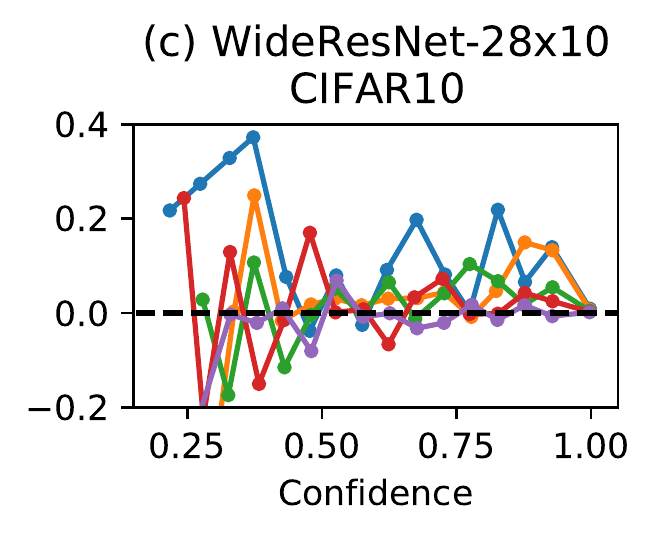}
     \end{subfigure}
     \hfill
     \begin{subfigure}[b]{0.324\textwidth}
         \centering
         \includegraphics[width=\textwidth]{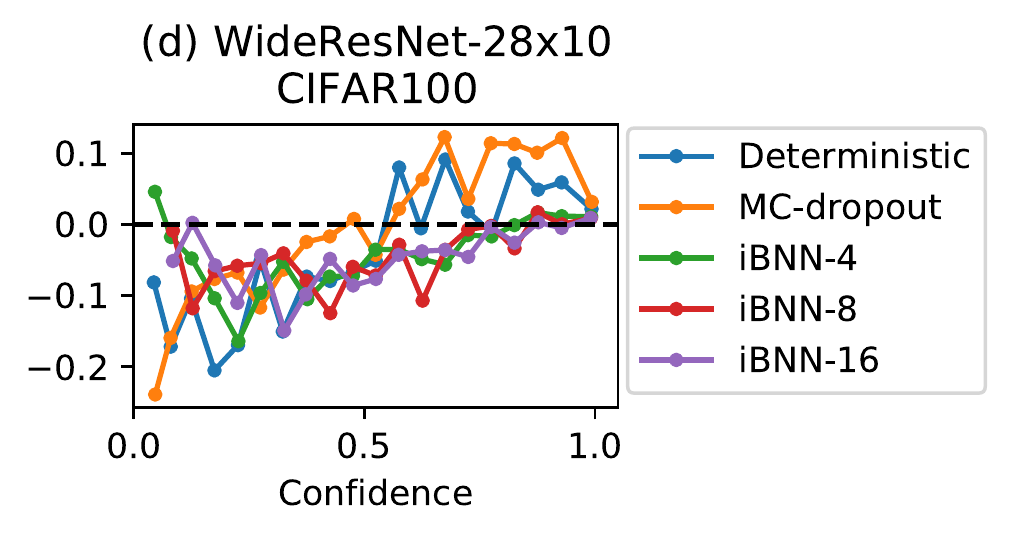}
     \end{subfigure}
    \caption{The calibration performance over \textsc{cifar} datasets with VGG-16 and WideResNet-28x10 networks. The iBNN is consistenly better calibrated than the competing methods.}
    \label{fig:calibration}
\end{figure*}

\begin{figure*}[t]
    \centering
    \begin{subfigure}[b]{0.226\textwidth}
         \centering
         \includegraphics[width=\textwidth]{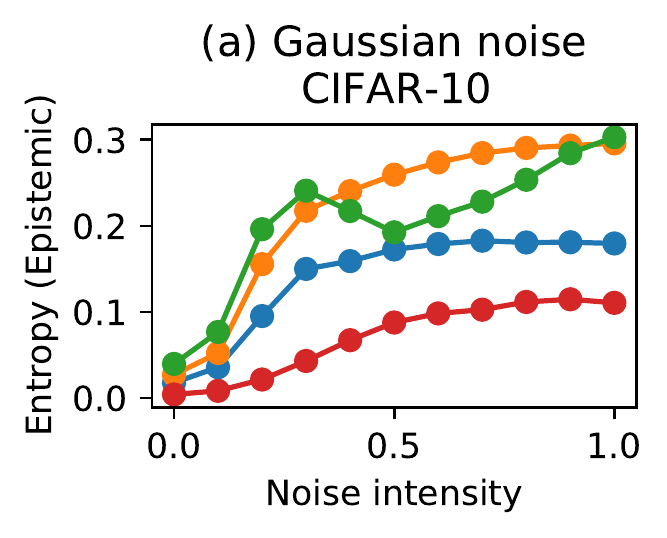}
     \end{subfigure}
     \hfill
     \begin{subfigure}[b]{0.209\textwidth}
         \centering
         \includegraphics[width=\textwidth]{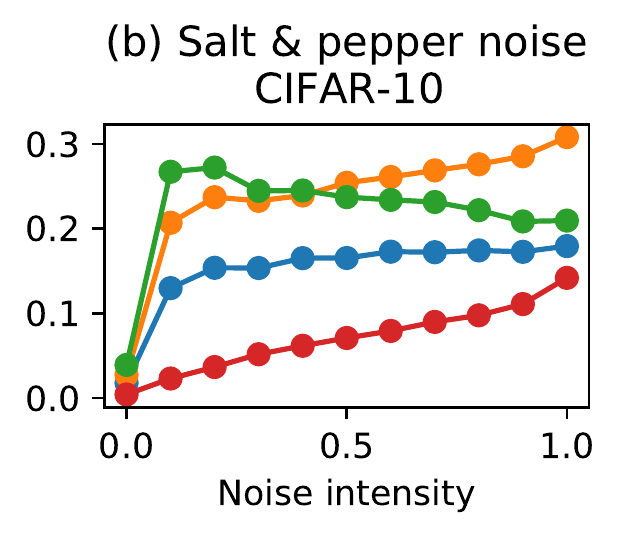}
     \end{subfigure}
     \hfill
     \begin{subfigure}[b]{0.209\textwidth}
         \centering
         \includegraphics[width=\textwidth]{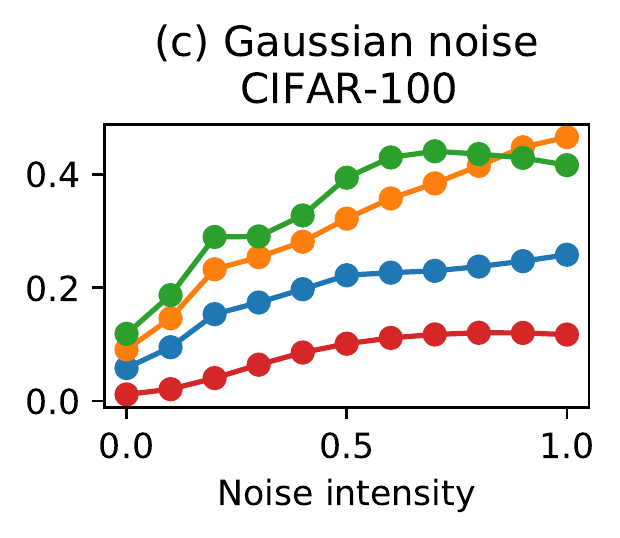}
     \end{subfigure}
     \hfill
     \begin{subfigure}[b]{0.329\textwidth}
         \centering
         \includegraphics[width=\textwidth]{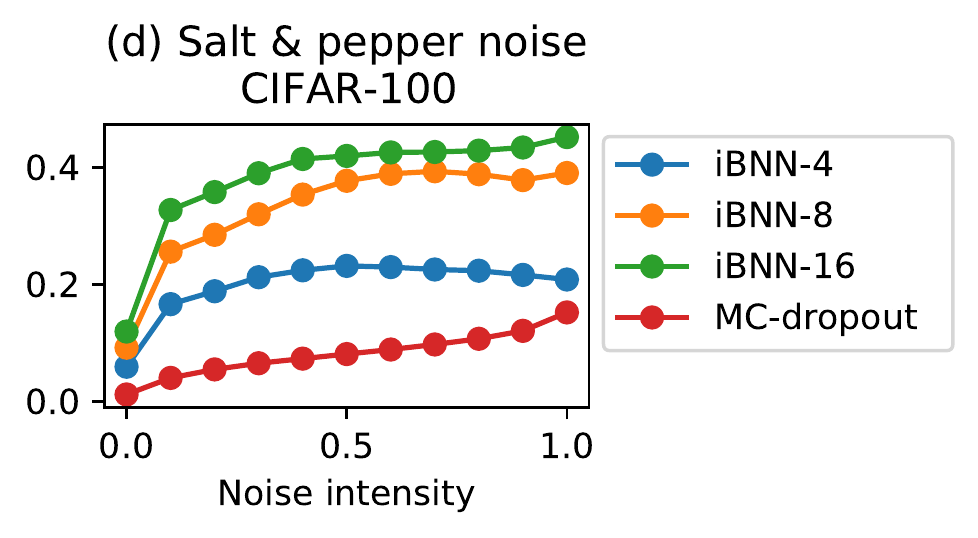}
     \end{subfigure}
     \begin{subfigure}[b]{0.226\textwidth}
         \centering
         \includegraphics[width=\textwidth]{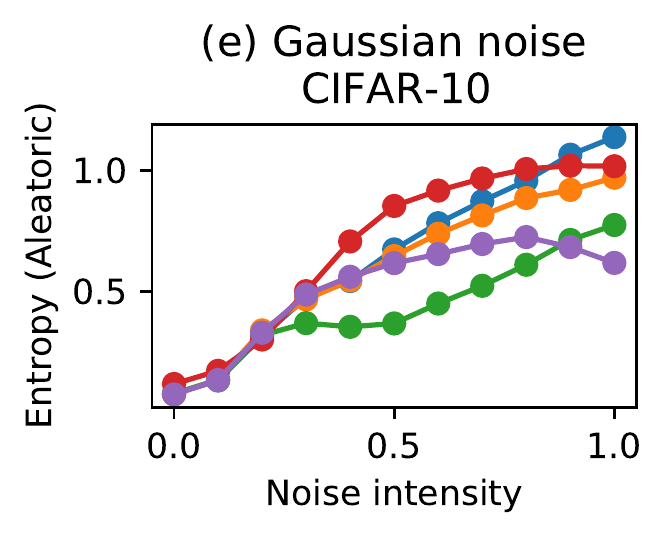}
     \end{subfigure}
     \hfill
     \begin{subfigure}[b]{0.209\textwidth}
         \centering
         \includegraphics[width=\textwidth]{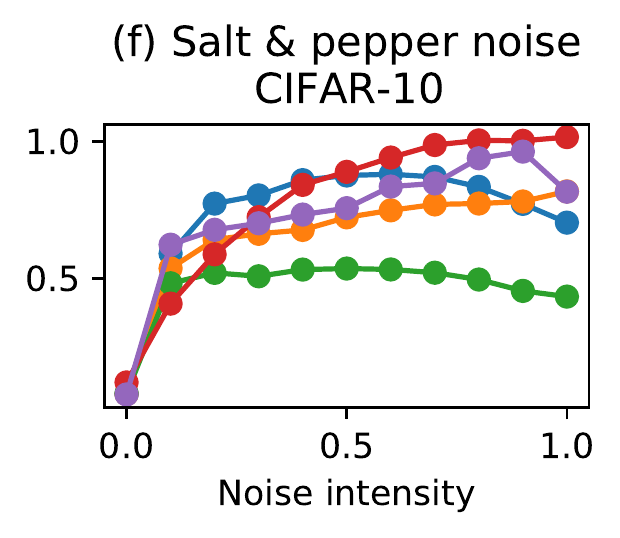}
     \end{subfigure}
     \hfill
     \begin{subfigure}[b]{0.209\textwidth}
         \centering
         \includegraphics[width=\textwidth]{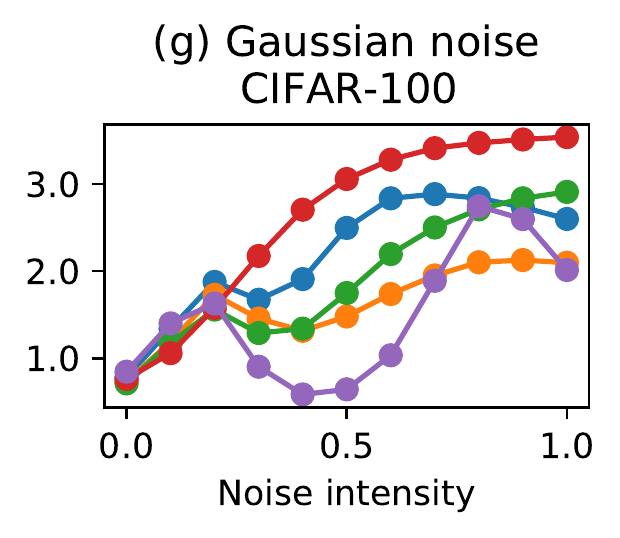}
     \end{subfigure}
     \hfill
     \begin{subfigure}[b]{0.329\textwidth}
         \centering
         \includegraphics[width=\textwidth]{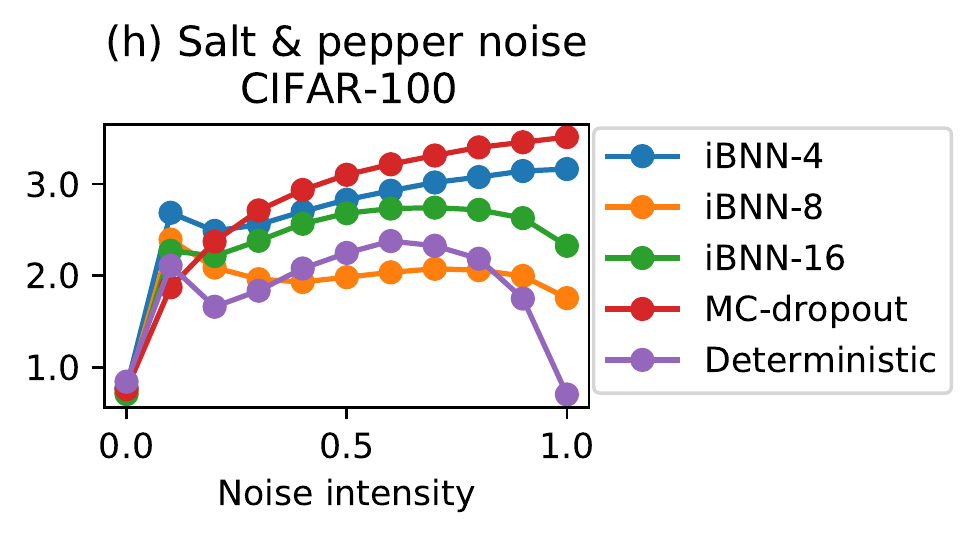}
     \end{subfigure}
    \caption{The mean predictive entropy (MPE) divided into aleatoric and epistemic parts over varying Gaussian or salt-and-pepper noise intensities. Both MC-dropout and iBNN correctly improve the predictive entropy with higher noise levels.}
    \label{fig:ood_epistemic}
\end{figure*}

The iBNN is significantly better than the deterministic model, the MC-dropout and BNN-VI methods across all metrics. The iBNN achieves the lowest predictive errors and negative log likelihoods, while also being the most calibrated model, according to the expected calibration error (ECE). We observe a general trend of improvement as we increase the number of components $K$ in the variational posterior, with iBNN-16 achieving almost consistenly the best performance.

We study the effect on the number of components more in-depth in Figure \ref{fig:effect_component}, where we demonstrate the performance of VGG-16 network on \textsc{cifar-100} with number of components $K \in \{4,8,16,24,32\}$. We notice that performance starts to decline when $K > 16$. Since during optimization each component in the posterior is optimized using a minibatch slice of size $B/K$ (see Algorithm \ref{algo:training}), for a fixed batch size $B$, a high number of components means a smaller slice, leading to more stochasticity in the optimization of each component, which seems to be harmful.

\subsection{Calibration}

Next, we study the calibration of the iBNN against its competing methods. Figure \ref{fig:calibration} demonstrates the discrepancy between prediction confidence (from softmax) and accuracy over the confidence range. A perfectly calibrated model achieves a zero discrepancy. The visualisation shows that the iBNN is consistently better calibrated than MC-dropout or deterministic neural networks. The improvement is especially significant in VGG-16 on \textsc{cifar-100}. Interestingly, the iBNN tends to slightly under-confident on \textsc{cifar-100} with WideResNet-28x10.

\subsection{Out-of-distribution performance}

We demonstrate the robustness of iBNN on out-of-distribution data by calculating the epistemic \eqref{eq:epi} and aleatoric \eqref{eq:ale} uncertainty of the WideResNet-28x10 model on corrupted \textsc{cifar-10} and \textsc{cifar-100} datasets. We create the out-of-distribution data by corrupting the samples with two methods:
\begin{enumerate}
    \item \textbf{Gaussian noise}: Each corrupted sample $\hat{\x}$ is created by mixing $\gamma \in [0,1]$ of standard Gaussian noise $\epsilon$ to $1-\gamma$ of the true sample $\x$:
    \begin{equation}
        \hat{\x} = (1-\gamma) \x + \gamma \epsilon, \qquad \epsilon \sim \N(\0, I).
    \end{equation}
    \item \textbf{Salt-and-pepper noise}: Each corrupted sample $\hat{\x}$ is created by randomly setting each element of $\x$ to 0 or 1 with probability $p$.
\end{enumerate}

Figure \ref{fig:ood_epistemic} show the change in the mean predictive entropy (MPE) of the WideResNet-28x10 model as we increase the Gaussian or salt-and-pepper noise intensity, with aleatoric (bottom) and epistemic (top) uncertainties shown. An ideal model would increase the predictive entropy as the input converges towards noise. Figure \ref{fig:ood_epistemic} shows that the iBNN model consistently increases the predictive entropy over increasingly noisy inputs. The iBNN tends to divide the uncertainty more towards epistemic component, while the MC-dropout tends to allocate the uncertainty more towards the aleatoric part.

% \subsection{Training time}
% Small to deep resnet on cifar-10, tracking accuracy and runtime.

\section{DISCUSSION}

In this paper we introduced implicit Bayesian neural networks that shift the inference of stochastic weights and biases to layer-wise input or hidden node variables. This shifts the task of inferring posterior over millions of weights into inferring a dramatically smaller posterior over thousands of nodes instead, while retaining the deep network capacity and representing weight stochasticity implicitly. We proposed an ensembled variational posterior inference, which scales efficiently to very large networks, such as VGG-16 with millions of parameters. We showed state-of-the-art performance in terms of prediction accuracy, calibration, out-of-distribution uncertainty characterisation, and robustness to noise. 

Even though we jointly optimize our deterministic weights and variational parameters in our main experiments, the formulation of the iBNN does permit direct intialisation using pre-trained weight parameters, while only learning the latent node parameters to represent multimodal uncertainty.  This notion can help reduce the training time significantly by splitting the procedure into two phases of optimizing the deterministic weights in the first phase, and jointly optimizing the both the weights and the variational parameters in the second phase. The inference time of iBNN is comparable to SGD or MC-dropout, since we can draw samples from the iBNN in parallel using modern GPUs.

We envision the input-wise prior paradigm to inspire several new research directions. The smaller posteriors open the possibility of more refined posterior inference via SG-MCMC \citep{wenzel2020good}, or network compression \citep{oneill2020}. The node prior perspective can provide a new view on network transformation equivariances and invariances \citep{finzi2020}. An immediate application of the presented work is in recurrent neural networks (RNNs) \citep{ma2019}, and especially in augmenting complex natural language processing models such as Transformers \citep{polosukhin2017} with Bayesian uncertainties. Finally, we also leave multiplicative bias priors as future work.

\subsubsection*{Acknowledgements}

This work was supported by the Academy of Finland (Flagship programme: Finnish Center for Artificial Intelligence FCAI, Grants 294238, 319264, 292334, 334600). We acknowledge the computational resources provided by Aalto Science-IT project, and the support from the Finnish Center for Artificial Intelligence (FCAI).

\bibliography{refs}

\begin{thebibliography}{27}
\providecommand{\natexlab}[1]{#1}
\providecommand{\url}[1]{\texttt{#1}}
\expandafter\ifx\csname urlstyle\endcsname\relax
  \providecommand{\doi}[1]{doi: #1}\else
  \providecommand{\doi}{doi: \begingroup \urlstyle{rm}\Url}\fi

\bibitem[Alemi et~al.(2018)Alemi, Poole, Fischer, Dillon, Saurous, and
  Murphy]{alemi2018}
Alexander Alemi, Ben Poole, Ian Fischer, Joshua Dillon, Rif~A Saurous, and
  Kevin Murphy.
\newblock Fixing a broken {ELBO}.
\newblock In \emph{ICML}, 2018.

\bibitem[Blei et~al.(2017)Blei, Kucukelbir, and McAuliffe]{blei2017}
David Blei, Alp Kucukelbir, and Jon McAuliffe.
\newblock Variational inference: A review for statisticians.
\newblock \emph{Journal of the American Statistical Association}, 112:\penalty0
  859--877, 2017.

\bibitem[Blundell et~al.(2015)Blundell, Cornebise, Kavukcuoglu, and
  Wierstra]{blundell2015weight}
Charles Blundell, Julien Cornebise, Koray Kavukcuoglu, and Daan Wierstra.
\newblock Weight uncertainty in neural networks.
\newblock In \emph{ICML}, 2015.

\bibitem[Depeweg et~al.(2018)Depeweg, Hern{\'a}ndez-Lobato, Doshi-Velez, and
  Udluft]{depeweg2017uncertainty}
Stefan Depeweg, Jos{\'e}~Miguel Hern{\'a}ndez-Lobato, Finale Doshi-Velez, and
  Steffen Udluft.
\newblock Uncertainty decomposition in {B}ayesian neural networks with latent
  variables.
\newblock In \emph{ICML}, 2018.

\bibitem[Finzi et~al.(2020)Finzi, Stanton, Izmailov, and Wilson]{finzi2020}
Marc Finzi, Samuel Stanton, Pavel Izmailov, and Andrew~Gordon Wilson.
\newblock Generalizing convolutional neural networks for equivariance to lie
  groups on arbitrary continuous data.
\newblock In \emph{ICML}, 2020.

\bibitem[Gal and Ghahramani(2016)]{gal2016dropout}
Yarin Gal and Zoubin Ghahramani.
\newblock Dropout as a {B}ayesian approximation: Representing model uncertainty
  in deep learning.
\newblock In \emph{ICML}, 2016.

\bibitem[Garipov et~al.(2018)Garipov, Izmailov, Podoprikhin, Vetrov, and
  Wilson]{garipov18}
Timur Garipov, Pavel Izmailov, Dmitrii Podoprikhin, Dmitry Vetrov, and
  Andrew~Gordon Wilson.
\newblock Loss surfaces, mode connectivity, and fast ensembling of dnns.
\newblock In \emph{NIPS}, 2018.

\bibitem[Guo et~al.(2017)Guo, Pleiss, Sun, and Weinberger]{guo2017}
Chuan Guo, Geoff Pleiss, Yu~Sun, and Kilian~Q. Weinberger.
\newblock On calibration of modern neural networks.
\newblock In \emph{ICML}, 2017.

\bibitem[Gustafsson et~al.(2020)Gustafsson, Danelljan, and
  Sch{\"o}n]{gustafsson2020}
Fredrik Gustafsson, Martin Danelljan, and Thomas Sch{\"o}n.
\newblock Evaluating scalable {B}ayesian deep learning methods for robust
  computer vision.
\newblock In \emph{CVPRW}, 2020.

\bibitem[Izmailov et~al.(2018)Izmailov, Podoprikhin, Garipov, Vetrov, and
  Wilson]{izmailov2018}
Pavel Izmailov, Dmitry Podoprikhin, Timur Garipov, Dmitry Vetrov, and
  Andrew~Gordon Wilson.
\newblock Averaging weights leads to wider optima and better generalization.
\newblock In \emph{UAI}, 2018.

\bibitem[Khan et~al.(2018)Khan, Nielsen, Tangkaratt, Lin, Gal, and
  Srivastava]{khan2018}
Mohammad~Emtiyaz Khan, Didrik Nielsen, Voot Tangkaratt, Wu~Lin, Yarin Gal, and
  Akash Srivastava.
\newblock Fast and scalable {B}ayesian deep learning by weight-perturbation in
  {A}dam.
\newblock In \emph{ICML}, 2018.

\bibitem[Lakshminarayanan et~al.(2017)Lakshminarayanan, Pritzel, and
  Blundell]{lakshminarayanan2017simple}
Balaji Lakshminarayanan, Alexander Pritzel, and Charles Blundell.
\newblock Simple and scalable predictive uncertainty estimation using deep
  ensembles.
\newblock In \emph{NIPS}, 2017.

\bibitem[Louizos and Welling(2017)]{louizos2017}
Christos Louizos and Max Welling.
\newblock Multiplicative normalizing flows for variational {B}ayesian neural
  networks.
\newblock In \emph{ICML}, 2017.

\bibitem[Ma et~al.(2019)Ma, Li, and Hernández-Lobato]{ma2019}
Chao Ma, Yingzhen Li, and José~Miguel Hernández-Lobato.
\newblock Variational implicit processes.
\newblock In \emph{ICML}, 2019.

\bibitem[Maddox et~al.(2019)Maddox, Izmailov, Garipov, Vetrov, and
  Wilson]{maddox2019simple}
Wesley~J Maddox, Pavel Izmailov, Timur Garipov, Dmitry~P Vetrov, and
  Andrew~Gordon Wilson.
\newblock A simple baseline for {B}ayesian uncertainty in deep learning.
\newblock In \emph{NIPS}, 2019.

\bibitem[O'Neill(2020)]{oneill2020}
James O'Neill.
\newblock An overview of neural network compression.
\newblock \emph{arXiv}, 2020.

\bibitem[Ovadia et~al.(2019)Ovadia, Fertig, Ren, Nado, Sculley, Nowozin,
  Dillon, Lakshminarayanan, and Snoek]{ovadia2019}
Yaniv Ovadia, Emily Fertig, Jie Ren, Zachary Nado, D.~Sculley, Sebastian
  Nowozin, Joshua Dillon, Balaji Lakshminarayanan, and Jasper Snoek.
\newblock Can you trust your model's uncertainty? {E}valuating predictive
  uncertainty under dataset shift.
\newblock In \emph{NIPS}, 2019.

\bibitem[Polosukhin et~al.(2017)Polosukhin, Kaiser, Gomez, Jones, Uszkoreit,
  Parmar, Shazeer, and Vaswani]{polosukhin2017}
Illia Polosukhin, Lukasz Kaiser, Aidan~N. Gomez, Llion Jones, Jakob Uszkoreit,
  Niki Parmar, Noam Shazeer, and Ashish Vaswani.
\newblock Attention is all you need.
\newblock In \emph{NIPS}, 2017.

\bibitem[Pradier et~al.(2018)Pradier, Pan, Yao, Ghosh, and
  Doshi-velez]{pradier2018}
Melanie Pradier, Weiwei Pan, Jiayu Yao, Soumya Ghosh, and Finale Doshi-velez.
\newblock Projected {BNN}s: {A}voiding weight-space pathologies by learning
  latent representations of neural network weights.
\newblock In \emph{NIPS workshop on {B}ayesian deep learning}, 2018.

\bibitem[Simonyan and Zisserman(2015)]{vgg}
Karen Simonyan and Andrew Zisserman.
\newblock Very deep convolutional networks for large-scale image recognition.
\newblock In \emph{ICLR}, 2015.

\bibitem[Su et~al.(2019)Su, Vargas, and Kouichi]{su2019}
Jiawei Su, Danilo~Vasconcellos Vargas, and Sakurai Kouichi.
\newblock One pixel attack for fooling deep neural networks.
\newblock \emph{IEEE Transactions on Evolutionary Computation}, 23\penalty0
  (5):\penalty0 828--841, 2019.

\bibitem[Sun et~al.(2019)Sun, Zhang, Shi, and Grosse]{sun2019functional}
Shengyang Sun, Guodong Zhang, Jiaxin Shi, and Roger Grosse.
\newblock Functional variational {B}ayesian neural networks.
\newblock In \emph{ICLR}, 2019.

\bibitem[Wenzel et~al.(2020)Wenzel, Roth, Veeling, Swiatkowski, Tran, Mandt,
  Snoek, Salimans, Jenatton, and Nowozin]{wenzel2020good}
Florian Wenzel, Kevin Roth, Bastiaan~S Veeling, Jakub Swiatkowski, Linh Tran,
  Stephan Mandt, Jasper Snoek, Tim Salimans, Rodolphe Jenatton, and Sebastian
  Nowozin.
\newblock How good is the {B}ayes posterior in deep neural networks really?
\newblock In \emph{ICML}, 2020.

\bibitem[Wilson(2020)]{wilson2020}
Andrew~Gordon Wilson.
\newblock The case for {B}ayesian deep learning.
\newblock \emph{arXiv:2001.10995}, 2020.

\bibitem[Zagoruyko and Komodakis(2016)]{wrn}
Sergey Zagoruyko and Nikos Komodakis.
\newblock Wide residual networks.
\newblock \emph{arXiv:1605.07146}, 2016.

\bibitem[Zhang et~al.(2017)Zhang, Bengio, Hardt, Recht, and Vinyals]{zhang2017}
Chiyuan Zhang, Samy Bengio, Moritz Hardt, Benjamin Recht, and Oriol Vinyals.
\newblock Understanding deep learning requires rethinking generalization.
\newblock In \emph{ICLR}, 2017.

\bibitem[Zhao et~al.(2020)Zhao, Liu, He, and Sun]{zhao2020}
Jing Zhao, Xiao Liu, Shaojie He, and Shiliang Sun.
\newblock Probabilistic inference of {B}ayesian neural networks with
  generalized expectation propagation.
\newblock \emph{Neurocomputing}, 412:\penalty0 392--398, 2020.

\end{thebibliography}

\clearpage
\onecolumn
\section*{Appendix}

\section*{Tables of test error, NLL and ECE}
We include the test error, negative log-likelihood (NLL) and expected calibration error (ECE) of the VGG-16 and WideResNet-28x10 models on \textsc{CIFAR-10} and \textsc{CIFAR-100} datasets in Table \ref{tab:acccifar}, \ref{tab:nllcifar} and \ref{tab:ececifar}. We calculate ECE using 15 bins. Each experiment is run 5 times with different random seeds. The iBNN greatly outperforms other methods in terms of test error and NLL, while producing more calibrated models according to the ECE. There is a clear trend that increasing the number of components in the posterior leads to better performances across all metrics. However, there is a limitation to this trend. We demonstrate this by training the VGG-16 on \textsc{CIFAR-100} dataset with various number of components $K \in \{4, 8, 16, 24, 32\}$ and report the result in Table \ref{tab:ncomponent_vgg16}. The result shows that the performance starts to decline when $K > 16$ in terms of NLL and ECE, and when $K > 24$ in terms of test error. We posit that due to our training algorithm, which trains the variational parameters of each component using a slice of size $B/K$ produced by dividing a mini-batch of size $B$ into $K$ equal parts, a larger $K$ leads to a smaller slice resulting in more noisy gradients for each component, which deteriorates the final performance. We believe that a larger number of components $K$ can be used without loss in performance if a larger batch size $B$ is employed during training.

\begin{table}[th]
    \caption{Error (lower is better) on \textsc{CIFAR-10} and \textsc{CIFAR-100}.}
    \label{tab:acccifar}
    \begin{center}
    \begin{tabular}{lcccc}
    \toprule
        & \multicolumn{2}{c}{CIFAR-10}        & \multicolumn{2}{c}{CIFAR-100}       \\ 
    \cmidrule(lr){2-3}     \cmidrule(lr){4-5}
               & VGG-16                    & WideResNet28x10           & VGG-16                    & WideResNet28x10  \\ \midrule
    SGD        & $6.72 \pm 0.06$           & $3.53 \pm 0.11$           & $26.74 \pm 0.13$          & $18.80 \pm 0.16$ \\
    MC-dropout & $6.71 \pm 0.12$           & $4.86 \pm 0.15$           & $26.41 \pm 0.20$          & $20.54 \pm 0.24$ \\
    BNN-VI     & $9.63 \pm 0.35$           & $5.96 \pm 0.28$           & $36.37 \pm 0.32$          & $24.13 \pm 0.40$ \\
    iBNN-4     & $6.33 \pm 0.14$           & $\mathbf{3.29 \pm 0.06}$  & $26.22 \pm 0.28$          & $17.79 \pm 0.24$ \\
    iBNN-8     & $6.31 \pm 0.08$           & $3.38 \pm 0.08$           & $25.89 \pm 0.19$          & $17.52 \pm 0.27$ \\
    iBNN-16    & $\mathbf{6.19 \pm 0.07}$  & $3.34 \pm 0.08$           & $\mathbf{25.72 \pm 0.08}$ & $\mathbf{17.27 \pm 0.09}$ \\ \bottomrule
    \end{tabular}
    \end{center}
\end{table}

\begin{table}[th]
    \caption{NLL (lower is better) on \textsc{CIFAR-10} and \textsc{CIFAR-100}.}
    \label{tab:nllcifar}
    \begin{center}
    \begin{tabular}{@{}lcccc@{}}
    \toprule
    & \multicolumn{2}{c}{CIFAR-10}              & \multicolumn{2}{c}{CIFAR-100}             \\     \cmidrule(lr){2-3}     \cmidrule(lr){4-5}
               & VGG-16                       & WideResNet28x10              & VGG-16                       & WideResNet28x10     \\ \midrule
    SGD        & $0.3255 \pm 0.0070$          & $0.1280 \pm 0.0014$          & $1.7232 \pm 0.0144$          & $0.7730 \pm 0.0078$ \\
    MC-dropout & $0.2887 \pm 0.0125$          & $0.1645 \pm 0.0082$          & $1.4048 \pm 0.0289$          & $0.8364 \pm 0.0104$ \\
    BNN-VI     & $0.6601 \pm 0.0358$          & $0.2685 \pm 0.0169$          & $2.3733 \pm 0.0954$          & $1.2683 \pm 0.0447$ \\
    iBNN-4     & $0.2147 \pm 0.0036$          & $0.1122 \pm 0.0011$          & $1.1249 \pm 0.0189$          & $0.6818 \pm 0.0055$ \\
    iBNN-8     & $0.2061 \pm 0.0031$          & $0.1093 \pm 0.0021$          & $1.0612 \pm 0.0074$          & $0.6601 \pm 0.0043$ \\
    iBNN-16    & $\mathbf{0.1983 \pm 0.0015}$ & $\mathbf{0.1073 \pm 0.0020}$ & $\mathbf{1.0325 \pm 0.0069}$ & $\mathbf{0.6440 \pm 0.0049}$ \\ \bottomrule
    \end{tabular}
    \end{center}
\end{table}

\begin{table}[th]
    \caption{ECE (lower is better) on \textsc{CIFAR-10} and \textsc{CIFAR-100}.}
    \label{tab:ececifar}
    \begin{center}
    \begin{tabular}{@{}lcccc@{}}
    \toprule
        & \multicolumn{2}{c}{CIFAR-10}              & \multicolumn{2}{c}{CIFAR-100}             \\    \cmidrule(lr){2-3}     \cmidrule(lr){4-5}
          & VGG-16                       & WideResNet28x10              & VGG-16                       & WideResNet28x10     \\ \midrule
    SGD        & $0.0476 \pm 0.0004$          & $0.0165 \pm 0.0003$          & $0.1869 \pm 0.0017$          & $0.0466 \pm 0.0015$ \\
    MC-dropout & $0.0421 \pm 0.0007$          & $0.0149 \pm 0.0009$          & $0.1457 \pm 0.0012$          & $0.0485 \pm 0.0020$ \\
    BNN-VI     & $0.0656 \pm 0.0015$          & $0.0290 \pm 0.0018$          & $0.1886 \pm 0.0016$          & $0.0909 \pm 0.0032$ \\
    iBNN-4     & $0.0173 \pm 0.0017$          & $0.0093 \pm 0.0007$          & $0.0807 \pm 0.0050$          & $0.0284 \pm 0.0021$ \\
    iBNN-8     & $0.0163 \pm 0.0010$          & $0.0064 \pm 0.0009$          & $0.0755 \pm 0.0018$          & $\mathbf{0.0274 \pm 0.0021}$ \\
    iBNN-16    & $\mathbf{0.0149 \pm 0.0008}$ & $\mathbf{0.0048 \pm 0.0013}$ & $\mathbf{0.0713 \pm 0.0028}$ & $0.0287 \pm 0.0014$ \\ \bottomrule
    \end{tabular}
    \end{center}
\end{table}

\begin{table}[ht]
\centering
\caption{Performance of VGG-16 model on \textsc{CIFAR-100} with different number of components in the posterior.}
\label{tab:ncomponent_vgg16}
\begin{tabular}{lccc}
\toprule
\# Components & Error ($\downarrow$)      & NLL ($\downarrow$)           & ECE ($\downarrow$) \\
\midrule
$K=4$         & $26.22 \pm 0.28$          & $1.1249 \pm 0.0189$          & $0.0807 \pm 0.0050$ \\
$K=8$         & $25.89 \pm 0.19$          & $1.0612 \pm 0.0074$          & $0.0755 \pm 0.0018$ \\
$K=16$        & $25.72 \pm 0.08$          & $\mathbf{1.0325 \pm 0.0069}$ & $\mathbf{0.0713 \pm 0.0028}$ \\
$K=24$        & $\mathbf{25.53 \pm 0.34}$ & $1.0467 \pm 0.0184$          & $0.0755 \pm 0.0035$ \\
$K=32$        & $25.67 \pm 0.23$          & $1.0758 \pm 0.0089$          & $0.0828 \pm 0.0019$ \\
\bottomrule
\end{tabular}

\end{table}

\section*{Using pretrained deterministic weights}
We demonstrate that our method could improve the performance of pretrained deterministic models. We use the pretrained ResNet-50 model on \textsc{ImageNet} from the \texttt{torchvision} package and run the training of the variational parameters for 15 epochs. We set the learning rate of the deterministic weights to a very small value of $0.0001$, and we set the learning rate of the variational parameters to $0.2$, which we anneal to $0.05$ for 4 epochs starting from epoch 6. We choose the prior to be $\mathcal{N}(1,0.1)$ and we use a batch size of 256. We report the result in Table \ref{tab:imagenet}. Our method provides a significantly improvement in performance across all metrics compared to the original pretrained solution. We want to note that this improvement depends a lot on the pretrained weights, since they decide the region on the loss surface that the variational parameters will explore.

\begin{table}[ht]
\centering
\caption{Performance of ResNet-50 on the validation set of \textsc{ImageNet} using pretrained parameters. Each experiment is run once.}
\label{tab:imagenet}
\begin{tabular}{lccc}
\toprule
 & Error ($\downarrow$)      & NLL ($\downarrow$)           & ECE ($\downarrow$) \\
\midrule
SGD            & $23.87$ & $0.9618$ & $0.0373$ \\
iBNN-4         & $23.22$          & $0.9310$          & $0.0350$ \\
iBNN-8         & $23.15$          & $0.9237$          & $0.0323$ \\
iBNN-16        & $\textbf{23.12}$          & $\mathbf{0.9118}$ & $\mathbf{0.0232}$ \\
\bottomrule
\end{tabular}
\end{table}

\end{document}